\documentclass{article} 
\pdfoutput=1
\usepackage{nips14submit_e,times}
\usepackage{hyperref}
\usepackage{url}

\usepackage{amsthm}
\usepackage{amsmath}
\usepackage{amssymb}

\usepackage{color}
\definecolor{red}{rgb}{0.88,0.22,0.21}

\usepackage{bbm}
\usepackage{graphicx}

\usepackage{algorithm}
\usepackage{algorithmic}

\usepackage{url}

\usepackage{pbox}

\usepackage{multirow}

\usepackage{rotating}

\title{Identifying Higher-order Combinations of Binary Features}

\author{
Felipe Llinares \\
D-BSSE, ETH Z\"{u}rich \\
\texttt{felipe.llinares@bsse.ethz.ch} \\
\and
\bf{Mahito Sugiyama} \\
ISIR, Osaka University\\
\texttt{mahito@ar.sanken.osaka-u.ac.jp} \\
\And
Karsten M. Borgwardt \\
D-BSSE, ETH Z\"{u}rich \\
\texttt{karsten.borgwardt@bsse.ethz.ch} \\
}

\nipsfinalcopy 

\begin{document}

\maketitle

\begin{abstract}
Finding statistically significant interactions between binary variables is computationally and statistically challenging in high-dimensional settings, due to the combinatorial explosion in the number of hypotheses.
Terada \textit{et al.}~recently showed how to elegantly address this multiple testing problem by excluding {\it non-testable hypotheses}. Still, it remains unclear how their approach scales to large datasets.

We here proposed strategies to speed up the approach by Terada et al. and evaluate them thoroughly in 11 real-world benchmark datasets. We observe that one approach, incremental search with early stopping, is orders of magnitude faster than the current state-of-the-art approach.

\end{abstract}

\section{Introduction}
\label{sec:Sec1}

The search for interactions between binary variables is of great importance in a diverse set of application domains. For instance in marketing \cite{Agrawal93}, one tries to find sets of products that are frequently co-bought by customers, and in genetics, one tries to find sets of mutations in the genome that frequently occur in disease carriers. 

A variety of approaches have been proposed for finding such {\it patterns} efficiently, that is interactions between binary variables, for instance in association rule mining  \cite{Agrawal93}, \cite{Toivonen} and recently Random Intersection Trees \cite{RIT}. Still, quantifying the statistical significance of the occurrence of a pattern was long deemed a hopeless endeavour in high-dimensional settings due to the combinatorial explosion of the number of possible patterns. In a dataset with $P$ variables, we could test up to $2^P$ possible variables. This in turn causes two main difficulties: (1) Naively listing all possible patterns is computationally unfeasible except for low-dimensional problems; and (2) the need to apply multiple testing correction with such a daunting number of patterns will almost surely result in an enormous loss of statistical power, that is the ability to find true patterns.

However, Terada et al.\ recently proposed a new algorithm for testing combinatorial interactions of binary predictors up to any order, the {\it Limitless Arity Multiple testing Procedure (LAMP)} \cite{Terada}. It builds upon work by Tarone \cite{Tarone}, who showed that ignoring {\it non-testable hypotheses} in multiple testing correction does not affect the family-wise error rate, but leads to an increase in statistical power. Terada et al.\ showed that this number of testable hypotheses can be computed via frequent item set mining.  
Even though the work in \cite{Terada} represents a big step forward, its procedure for finding the number of testable hypotheses is slow on large datasets, with many objects $N$.
Their search procedure iteratively searches all frequent item sets with a frequency of $\sigma$, decrementing $\sigma$ in each iteration. If $N$ is large, the range of admissible values for $\sigma$ is also large and results in a longer runtime.

{\it Here we address exactly this problem by empirically evaluating strategies to speed up LAMP. We observe that one of these strategies is on 10 out of 11 datasets at least one order of magnitude faster, often even several orders of magnitude faster than the original LAMP, while returning the exact solution.}

The organisation of the paper is as follows: In Section \ref{sec:Sec2} we define the problem of testing high-order interactions between binary predictors. In Section 3 we review the improved Bonferroni correction proposed in \cite{Tarone}, summarize the original version of LAMP and present the speed-up procedures we propose. Section 4 demonstrates empirically the speed improvement achieved by our scheme. Finally, Section 5 sums up the main conclusions of this work.
 



\section{Testing the statistical significance of combinatorial interactions}
\label{sec:Sec2}

Let $X \in \left\{0,1\right\}^{N \times P}$ be a dataset consisting of $P$ binary features and $N$ observations and $y \in \left\{0,1\right\}^{N}$ be the corresponding vector of binary labels. For any subset $S \subset \left\{1,2,\hdots,P\right\}$ of $|S|$ features, we define the \emph{high-order interaction} among the $|S|$ features contained in $S$ as $X_{i,S} = \prod_{j \in S}X_{ij} \; \forall \; i=1,\hdots,N$. Excluding the empty set, we can construct up to $2^{P} - 1$ such high-order interaction features, which will capture interactions of different orders depending on the cardinality of $S$. Individual features are included in this formulation too as all singleton sets $S : |S|=1$.

Then, the problem of testing the association of all possible combinations of $P$ binary predictors with a dichotomous label can be naively formulated as a multiple hypothesis testing problem consisting of $2^{P} - 1$ parallel univariate association tests between two binary random variables. Those are usually represented in terms of 2x2 contingency tables as the one depicted in Table \ref{tab:contingency_table}.

\begin{table}[ht]
\centering
	\begin{tabular}{|c|c|c|c|}
		\hline
		Variables & $X_{i,S}=1$ & $X_{i,S}=0$ & Row marginals\\
		\hline
		$y_{i}=1$ & $a_{S}$ & $n-a_{S}$ & $n$ \\
		\hline
		$y_{i}=0$ & $x_{S}-a_{S}$ & $N+a_{S}-n-x_{S}$ & $N-n$ \\
		\hline
		Col marginals & $x_{S}$ & $N-x_{S}$ & $N$ \\
		\hline
	\end{tabular}
\caption{2x2 contingency table for testing the association of an arbitrary high-order interaction feature $X_{i,S}$ with the class labels $y_{i}$}
\label{tab:contingency_table}
\end{table}

One of the most popular statistical association tests for 2x2 contingency tables is Fisher's exact test \cite{FisherExactTest}. Under the null hypothesis of no association between the two random variables, it can be shown that $p(a_{S} | x_{S},n,N)$ follows a Hypergeometric distribution\footnote{$p(a_{S} | x_{S},n,N)= \binom{n}{a_{S}}\binom{N-n}{x_{S}-a_{S}} \Big/ \binom{N}{x_{S}}$}. The probability of observing a table at least as extreme as the one we actually observed, that is, the \emph{P-value} of the table, can be then computed as $P_{\mathrm{val}}^{(S)} = \sum_{k=a_{S}}^{\mathrm{min}(x_{S},n)}{p(k | x_{S},n,N)}$. That corresponds to a \emph{one-tailed test}. If a \emph{two-tailed test} is desired, one simple solution is to compute the one-sided P-value of the smallest tail and then double it to account for the other tail \cite{bland2000}.

We say that a high-order interaction $S$ is \emph{significant} at level $\alpha$ if the P-value is smaller than $\alpha$. By construction, the probability of deeming an association significant when the null hypothesis of no association is true is upper bounded by $\alpha$. However, if $m$ association tests are run in parallel, the probability that at least one of those $m$ tests will result in a false positive will be much greater than the original uncorrected significance level $\alpha$. Bounding that probability, denoted Family Wise Error Rate (FWER), is the main goal of many multiple hypothesis testing procedures. 

One of the most common schemes to do so is the Bonferroni correction \cite{Bonferroni36}. The idea is simple: applying the union bound we obtain $\mathrm{FWER} \leq \alpha m$. This shows that using a \emph{corrected significance threshold} $\delta=\alpha / m$ rather than the original \emph{uncorrected significance threshold} $\alpha$ for each individual association test yields a multiple hypothesis testing procedure satisfying $\mathrm{FWER} \leq \alpha$. Since the bound $\mathrm{FWER} \leq \alpha m$ is only tight under assumption that all $m$ tests are jointly independent, the Bonferroni correction tends to be overly conservative, causing a loss of statistical power.  

\section{Speeding up the search for significant interactions}
\label{sec:Sec3}

The combinatorial explosion of the number of hypotheses generated when testing for higher-order interactions poses a real challenge for the application of a Bonferroni-style correction. Since the \emph{Bonferroni correction factor} grows exponentially with the number of features, $m=2^{P} - 1$, only exceptionally strong association signals will be deemed significant after correcting the significance threshold.

This has acted as a major discouraging factor, hindering the development of statistical testing approaches for high-order interactions of features. To the best of our knowledge, \cite{Terada} has been the first work to come up with a feasible scheme to carry out statistical significance testing of combinatorial interactions of features, focusing on the context of gene regulatory motif discovery. Their work is based on the idea of the ``testability of hypotheses'' from \cite{Tarone}.

\subsection{Refined Bonferroni correction for discrete test statistics}
\label{sec:Sec3p1}

The main idea behind \cite{Tarone} is that in a multiple hypothesis testing problem involving binary random variables, a potentially large number of tests cannot possibly be significant irrespectively of the actual observed cell counts and, therefore, do not need to be neither tested nor taken into account when computing the Bonferroni correction factor. This can lead to a big reduction in both the computational burden and the loss of statistical power implied by testing a large number of hypotheses in parallel.

There are two main reasons why such a procedure can be carried out: (1) if the data is discrete, the test statistic can only attain a finite set of values, hence a minimum attainable P-value will exist; and (2) in some cases, most notably Fisher's exact test for 2x2 contingency tables, such minimum attainable P-value can be easily computed and depends only on the marginals $x_{S}$, $n$ and $N$. In this way, if the minimum attainable P-value $\Psi(x_{S},n,N)$ of the association test for feature subset $S$ is above the corrected significance threshold $\delta$, there is no need to inspect the inner cell counts of the table $a_{S}$ and compute the corresponding P-value as the test cannot possibly yield a significant result. We call hypothesis tests for which $\Psi(x_{S},n,N) \le \delta$ \emph{testable hypotheses}.

It is shown in \cite[Supporting Text 4]{Terada} that, under the assumption\footnote{The assumption $n<N-n$ does not imply a loss of generality, since the labels can be swapped. The assumption $x_{S} \le n$ is reasonable for most real-world datasets. If $x_{S} > n$, then $\Psi(x_{S},n,N)$ is no longer non-increasing on $x_{S}$. However, a simply workaround is proposed in \cite[Supporting Text 4]{Terada} by defining $\Psi(x_{S},n,N) = 1 / \binom{N}{n}$ in that scenario. With that definition, $\Psi(x_{S},n,N)$ is non-increasing in $x_{S}$, since it is constant, and is a lower bound in the real minimum attainable P-value.} that $n<N-n$ and $x_{S} \le n$, the minimum P-value attainable by Fisher's exact test is given by $\Psi(x_{S},n,N)=\binom{n}{x_{S}} / \binom{N}{x_{S}}$. A crucial remark is that $\Psi(x_{S},n,N)$ is a non-increasing function of $x_{S}$ when the aforementioned assumptions hold.

Observation (2) above, which is in our opinion not emphasized enough in \cite{Tarone}, is fundamental to understand why this test-discarding scheme is actually valid. Since the null distribution in Fisher's exact test is conditioned on the marginals $x_{S}$, $n$ and $N$ of the corresponding 2x2 contingency table, we have that $p(a_{S} | x_{S},n,N,\Psi(x_{S},n,N)) = p(a_{S} | x_{S},n,N)$. Therefore, the null distribution of the test statistics is not modified by keeping only those tests whose minimum attainable P-value is smaller than $\delta$.

As the corrected significance threshold $\delta$ depends on the number of testable hypotheses and the testability status of a hypothesis depends on $\delta$, determining both $\delta$ and the set of testable hypotheses can be treated as finding the \emph{rounded root} of a function $f(k)=m(k)-k$ defined for natural numbers $ k \in \mathbb{N}$. Let $\mathcal{S}$ be the set of all possible subsets of features $S$. Let $m(k)$ be the number of hypotheses which are testable at level $\delta=\alpha / k$, that is, $m(k) := \left| \left\{S \in \mathcal{S} | \Psi(x_{S},n,N) \le \alpha / k  \right\} \right|$. Then, if we find $k_{\mathrm{rt}}$ such that $m(k_{\mathrm{rt}}-1) > k_{\mathrm{rt}}-1$ and $m(k_{\mathrm{rt}}) \le k_{\mathrm{rt}}$, that is, the rounded root of $m(k)-k$, we have that the set of testable hypotheses will be  $\left| \left\{S \in \mathcal{S} | \Psi(x_{S},n,N) \le \alpha / k_{\mathrm{rt}}  \right\} \right|$ and using $m(k_{\mathrm{rt}})$ as Bonferroni correction factor guarantees $FWER \le \alpha$.

\subsection{Efficient enumeration of testable patterns}
\label{sec:Sec3p2}

In \cite{Tarone}, the refined Bonferroni correction factor is discussed as a general tool applicable to any multiple hypothesis testing problem involving discrete test statistics. However, the author does not address the issue of how to efficiently find $k_{\mathrm{rt}}$. In fact, in the context of searching for significant higher-order combinations of features, naively evaluating $m(k)$ would require listing all $2^{P}-1$ possible combinations of features $S \in \mathcal{S}$ and computing the minimum P-value $\Psi(x_{S},n,N)$ of every single combination. The computational cost would therefore be prohibitive even for small datasets.

The Limitless-Arity Multiple testing Procedure (LAMP) described in \cite{Terada} circumvents that problem by establishing a connection between enumeration of all testable patterns and the well studied problem of \emph{frequent itemset mining}.

The key observation is that the minimum attainable P-value $\Psi(x_{S},n,N)$ is a non-increasing function of $x_{S}$. This implies that there is a one-to-one mapping between $k_{\mathrm{rt}}$ and the \emph{root frequency} $\sigma_{\mathrm{rt}}$ defined as the natural number which satisfies $\left| \left\{S \in \mathcal{S} | x_{S} \ge \sigma_{\mathrm{rt}}-1 \right\} \right| > \alpha / \Psi(\sigma_{\mathrm{rt}}-1,n,N)$ and $\left| \left\{S \in \mathcal{S} | x_{S} \ge \sigma_{\mathrm{rt}} \right\} \right| \le \alpha / \Psi(\sigma_{\mathrm{rt}},n,N)$. This can be shown precisely by letting $k_{\mathrm{rt}}=\alpha / \Psi(\sigma_{\mathrm{rt}},n,N)$ and recalling the definition  $m(k) := \left| \left\{S \in \mathcal{S} | \Psi(x_{S},n,N) \le \alpha / k  \right\} \right|$. Then $m(k_{\mathrm{rt}}) = m(\alpha / \Psi(\sigma_{\mathrm{rt}},n,N)) = \left| \left\{S \in \mathcal{S} | \Psi(x_{S}) \le \Psi(\sigma_{\mathrm{rt}},n,N)  \right\} \right| = \left| \left\{S \in \mathcal{S} | x_{S} \ge \sigma_{\mathrm{rt}} \right\} \right|$. That proves that the set of testable hypotheses coincides with the set of patterns for which $x_{S} \ge \sigma_{\mathrm{rt}}$. 

Let us consider that each of the $P$ features is an \emph{item}, and that an observation in the dataset $X$ is a \emph{transaction} that contains those items corresponding to features taking value 1 in the observation. In this context, the $N$ transactions corresponding to the $N$ observations in the dataset $X$ form a \emph{transactions database}.

Then, the problem of finding the set $\left\{S \in \mathcal{S} | x_{S} \ge \sigma \right\}$ for a given value of $\sigma$ is equivalent to finding all combinations of items or \emph{itemsets} which appear in at least $\sigma$ transactions in the transactions database. We say that those itemsets are \emph{frequent} with \emph{support} $\sigma$.

This dual formulation allows using off-the-shelf frequent itemset mining algorithm in order to efficiently enumerate the set $\left\{S \in \mathcal{S} | x_{S} \ge \sigma \right\}$ for different values of $\sigma$. Those can be combined with an appropriate root searching procedure in order to find the root frequency $\sigma_{\mathrm{rt}}$ ,which in turn determines $m(k_{\mathrm{rt}})$ and the set of testable tests as shown above.

In the original LAMP formulation described in \cite{Terada}, a decremental search procedure is employed to find $\sigma_{\mathrm{rt}}$. The algorithm begins by initializing $\sigma=n$. Then, while $\left| \left\{S \in \mathcal{S} | x_{S} \ge \sigma \right\} \right| \le \alpha / \Psi(\sigma,n,N)$, $\sigma$ is iteratively decreased in steps of 1. Once a value of $\sigma$ that satisfies $\left| \left\{S \in \mathcal{S} | x_{S} \ge \sigma \right\} \right| > \alpha / \Psi(\sigma,n,N)$ is found, we know that $\sigma_{\mathrm{rt}}=\sigma+1$ and the algorithm terminates. The pseudocode is summarized in Algorithm \ref{alg:LAMP}.

\begin{algorithm}
\caption{LAMP \label{alg:LAMP}}
\begin{algorithmic}[1]
	\STATE  {\bfseries Input:} $X$, $y$, $n$, $N$
	\STATE  {\bfseries Output:} All significant high-order interactions among features in $X$
	\STATE  $\sigma \leftarrow n+1$
	\REPEAT
    		\STATE $\sigma \leftarrow \sigma-1$
		\STATE Enumerate all frequent itemsets with support $\sigma$
		\STATE $\mathcal{T}(\sigma) \leftarrow \left\{S \in \mathcal{S} | x_{S} \ge \sigma \right\}$
        \UNTIL{$\left|\mathcal{T}(\sigma)\right| > \alpha / \Psi(\sigma,n,N)$}
    \STATE $\sigma_{\mathrm{rt}} \leftarrow \sigma + 1$
    \STATE $\mathcal{T}(\sigma_{\mathrm{rt}}) \leftarrow \left\{S \in \mathcal{S} | x_{S} \ge \sigma_{\mathrm{rt}} \right\}$
    \STATE $\mathcal{S}_{\mathrm{sig}} \leftarrow \left\{ S \in \mathcal{T}(\sigma_{\mathrm{rt}}) | P_{\mathrm{val}}^{(S)} \leq \alpha / \left| \mathcal{T}(\sigma_{\mathrm{rt}})  \right| \right\}$
\STATE {\bfseries Return:} $\mathcal{S}_{\mathrm{sig}}$
\end{algorithmic}
\end{algorithm}

\subsection{Speeding up LAMP}

While LAMP represents the first computationally feasible algorithm able to mine statistically significant interactions between binary features, it can still be too slow even for medium-sized datasets. 

In what follows, we describe two non-exclusive approaches which can be used to reduce the execution time of LAMP: (1) an incremental search scheme with early stopping instead than the decremental search procedure used in \cite{Terada} and; (2) using a subsampled version $X^{\prime}$ of the original dataset $X$ in order to obtain a cheap estimation of the root frequency.

\subsubsection{Incremental search with early stopping}

Running a frequent itemset miner with $\sigma=1$ would list every combination of features which occurs at least once in the dataset, which in turn would allow to compute the root frequency $\sigma_{\mathrm{rt}}$ with a single execution of the miner. However, this approach would generate an enormous amount of untestable patterns, unnecessarily increasing the computation time considerably. The computational complexity of frequent itemset mining is proportional to the number of frequent itemsets, which rapidly increases as the support $\sigma$ is reduced. It seems then reasonable to initialize $\sigma$ to the largest possible value that $\sigma_{\mathrm{rt}}$ could take and iteratively decrement it until the solution is found. That is exactly what the original version of LAMP proposes. However, we believe that such reasoning ignores two subtle facts.

Since the minimum attainable P-value $\Psi(\sigma,n,N)$ decays super-exponentially with $\sigma$, those datasets for which $\sigma_{\mathrm{rt}}$ ends up being large will have an enormous number of testable patterns. In those cases, the modified Bonferroni correction works in a regime virtually identical to the naive Bonferroni correction and the search for high-order interactions via statistical testing will most likely fail regardless of which scheme is employed. This suggests that we should optimize our search scheme for datasets with a reasonably small $\sigma_{\mathrm{rt}}$, in which case an incremental search strategy makes more sense since it will require a smaller number of iterations to find the root frequency.

Even more importantly, while running the frequent itemset miner for small values of $\sigma$ could be slow, it is not necessary to enumerate all frequent itemsets. Once the number of frequent itemsets found is larger than $\alpha / \Psi(\sigma,n,N)$, we know that the root frequency $\sigma_{\mathrm{rt}}$ must be larger than the current value of $\sigma$ so we can stop the frequent itemset miner early. This effectively makes each iteration of incremental search no more computationally intensive than those of decremental search.

All in all, we claim that for real-world datasets for which the statistical assessment of higher-order feature interactions is feasible, an incremental search scheme will require less executions of the frequent itemset miner and early stopping will make each of those executions just as efficient as those of decremental search would be, which should result in a net speedup over the original implementation of LAMP. The pseudocode of incremental LAMP with early stopping is summarized in Algorithm \ref{alg:incLAMP}.

\begin{algorithm}
\caption{Incremental LAMP with early stopping \label{alg:incLAMP}}
\begin{algorithmic}[1]
	\STATE  {\bfseries Input:} $X$, $y$, $n$, $N$
	\STATE  {\bfseries Output:}  All significant high-order interactions among features in $X$
	\STATE $\sigma \leftarrow 0$
	\REPEAT
    		\STATE $\sigma \leftarrow \sigma+1$
		\STATE Enumerate frequent itemsets with support $\sigma$ recording the number m of frequent itemsets found so far
		\IF {$m > \alpha / \Psi(\sigma,n,N)$}
			\STATE   {\bfseries continue}
		\ELSE
			\STATE  $\sigma_{\mathrm{rt}} \leftarrow \sigma$
			\STATE $\mathcal{T}(\sigma_{\mathrm{rt}}) \leftarrow \left\{S \in \mathcal{S} | x_{S} \ge \sigma \right\}$
			\STATE  {\bfseries break}
		\ENDIF
        \UNTIL{false}
    \STATE $\mathcal{S}_{\mathrm{sig}} \leftarrow \left\{ S \in \mathcal{T}(\sigma_{\mathrm{rt}}) | P_{\mathrm{val}}^{(S)} \leq \alpha / \left| \mathcal{T}(\sigma_{\mathrm{rt}})  \right| \right\}$
    \STATE {\bfseries Return:} $\mathcal{S}_{\mathrm{sig}}$
\end{algorithmic}
\end{algorithm}

\subsubsection{Subsampling}

We also explore the possibility of applying subsampling to speed up LAMP by subsampling, a strategy proposed for association rule mining by \cite{Toivonen}. Suppose we construct a subsampled dataset $X^{\prime}$ by sampling with replacement $N/K$ observations from the original dataset $X$. Let us denote the number of occurrences of an arbitrary high-order interaction of features in the subsampled dataset by $x_{S}^{\prime}$. It is straightforward to show that $x_{S}^{\prime} \sim \mathrm{Binomial}(N/K,x_{S}/N)$, so that $\mathbb{E}[x_{S}^{\prime}]=x_{S}/K$. 

Roughly, subsampling the original dataset by a factor of $K$ scales the number of occurrences of every pattern by a factor of $1/K$. This suggests estimating the root frequency of the original dataset $X$, $\sigma_{\mathrm{rt}}$, from the occurrence counts of patterns in the subsampled dataset $X^{\prime}$ by finding $\sigma_{\mathrm{rt}}^{\prime}$
such that $\left| \left\{S \in \mathcal{S} | x_{S} \ge \sigma_{\mathrm{rt}}^{\prime}-1 \right\} \right| > \alpha / \Psi(K(\sigma_{\mathrm{rt}}^{\prime}-1),n,N)$ and $\left| \left\{S \in \mathcal{S} | x_{S} \ge \sigma_{\mathrm{rt}}^{\prime} \right\} \right| \le \alpha / \Psi(K\sigma_{\mathrm{rt}}^{\prime},n,N)$ and letting $\hat{\sigma}_{\mathrm{rt}}=K\sigma_{\mathrm{rt}}^{\prime}$. Note that the resolution in the determination of the original $\sigma_{\mathrm{rt}}$ from $\sigma_{\mathrm{rt}}^{\prime}$ is decimated by a factor of $K$.

\section{Experiments}

We evaluate the performance of our proposed methods compared with the original version of LAMP \cite{Terada} using 11 different real-world datasets. In Table~\ref{table:datasets_summary} we depict the most salient properties of each dataset: the number of samples $N$, the number of features $P$ and the root frequency $\sigma_{\mathrm{rt}}$ computed with LAMP for a target FWER of $0.05$. For the 4 labeled datasets, we use the true number of positive samples $n$ when enumerating testable hypotheses. For the remaining 7 datasets, for which there are no labels available, we examined two different scenarios\footnote{Note that the computational complexity of both the original LAMP and our proposals depends on the labels only through the ratio $n/N$ since the minimum attainable P-value $\Psi(x_{S},n,N) \approx (n/N)^{x_{S}}$}: $n=N/2$ and $n=N/10$, denoted $r=2$ and $r=10$ respectively. Thus 18 different cases are examined in our experiments. A more detailed description of the datasets can be found in the supplementary information.

All experiments were run on a server running Ubuntu 12.04.3 on a single core of an AMD Opteron CPU clocked at 2.6 GHz. The frequent itemset miner employed in all experiments was LCM \cite{Uno04anefficient} version 3, which has been shown to exhibit state-of-the-art performance in a great number of datasets and won the FIMI'04 frequent itemset mining competition \cite{FIMIDatasets}. The code of LCM is written in C and was compiled using \texttt{gcc 4.6.3} with \texttt{-O3} and \texttt{-march=opteron} as flags.

\begin{table}[t]
	\caption{Dataset characteristics} 
 \centering 
\begin{small}
	\begin{tabular}{l r r r r r r r r r r r} 
		\hline
		& \rotatebox[origin=c]{90}{\small{Tic-tac-toe}} & \rotatebox[origin=c]{90}{Chess} & \rotatebox[origin=c]{90}{Inetads} & \rotatebox[origin=c]{90}{Mushroom} & \rotatebox[origin=c]{90}{RCV1} & \rotatebox[origin=c]{90}{Connect} & \small{\rotatebox[origin=c]{90}{BMS-Web2}} & \rotatebox[origin=c]{90}{retail} & \rotatebox[origin=c]{90}{\small{T10I4D100K}} & \rotatebox[origin=c]{90}{\small T40I10D100K} & \rotatebox[origin=c]{90}{BMS-POS}\\
		\hline 
		$N$ & 958 & 3,196 & 3,279 & 8,124 & 20,242 & 67,557 & 77,512 & 88,162 & $10^5$ & $10^5$ & 515,597\\ 
		$P$ & 18 & 75 & 1,554 & 117 & 44,504 & 129 & 3,340 & 16,470 & 870 & 870 & 1,657\\%
		$\sigma_{\mathrm{rt}}$ & 11  & ---    & 13 & 31  & 41 & ---   & ---   & ---   & ---   & ---   & ---\\%
		$\sigma_{\mathrm{rt},r=2}$ & ---    & 44 & ---   & ---     & ---    & 55 & 23 & 21 & 22 & 34 & 32\\%
		$\sigma_{\mathrm{rt},r=10}$ & ---    & 16 & ---   & ---     & ---    & 18 & 9 & 7 & 8 & 12 & 11\\%
		\hline 
	\end{tabular}
 \end{small}
\label{table:datasets_summary} 
\end{table} 

\begin{figure}[t]
	\begin{center}
		\includegraphics[width=1.0\textwidth]{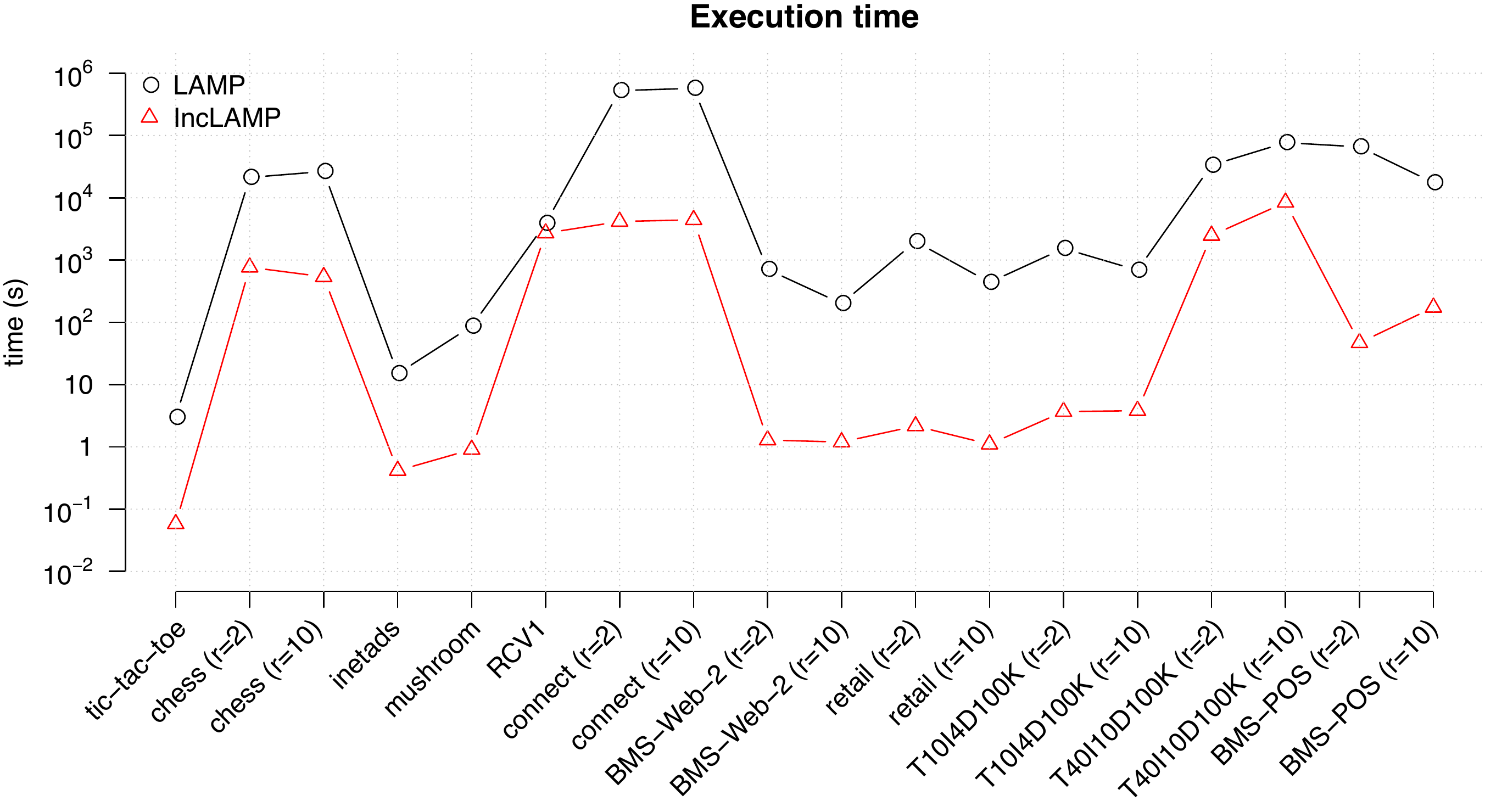}
	\end{center}
	\caption{Comparison in terms of runtime between the original version of LAMP in \cite{Terada} and our proposal using incremental search with early stopping (IncLAMP)}
	\label{fig:incLAMPvsLAMP}
\end{figure}

\begin{figure}[t]
	\begin{center}
	 \includegraphics[width=1.0\textwidth]{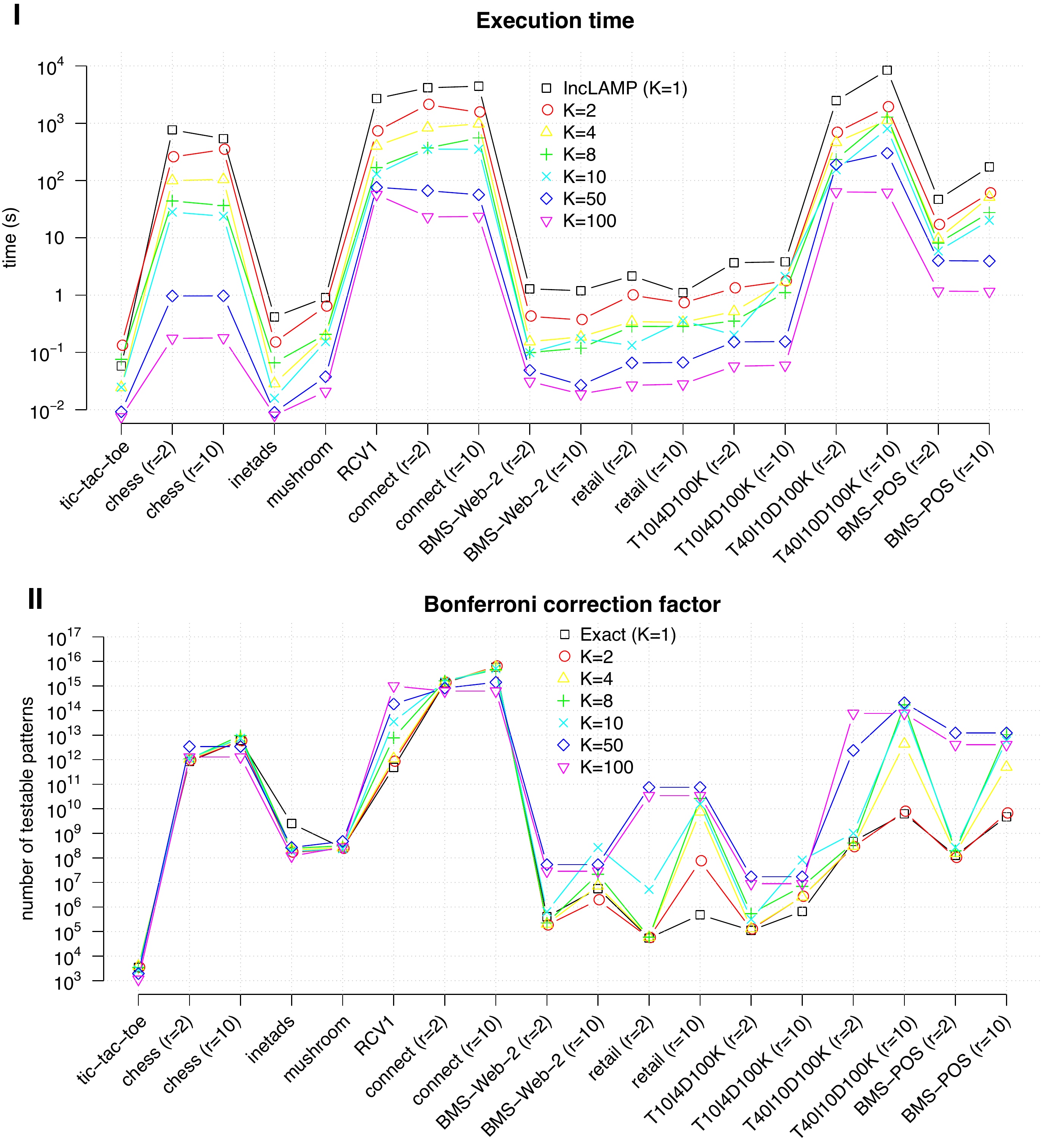}
	\end{center}
	\caption{(I) Effect of subsampling ratio on the execution time for IncLAMP (II) Estimated number of testable patterns for different subsampling ratios}
	\label{fig:subsampling}
\end{figure}

Figure~\ref{fig:incLAMPvsLAMP} shows the runtime of the original implementation of LAMP versus our scheme based on incremental search and early stopping. The speedup is dramatic: our approach is one order of magnitude faster than the original LAMP in 6 out of 18 cases, two orders of magnitude faster in 9 out of 18 cases and three orders in 2 out of 18 cases. Only one dataset in the whole set of 18 cases shows a modest speedup, and even in that case we achieve a speedup of roughly 50\%.

The computation time depends on a non-trivial way in the characteristics of datasets. Apart from the number of samples and the number of features, the distribution of frequent patterns with respect to changes in the support $\sigma$ and the resulting root frequency $\sigma_{\mathrm{rt}}$ are critical. The computational complexity of each execution of frequent itemset mining depends on the number of frequent patterns $|\left\{S \in \mathcal{S} | x_{S} \ge \sigma \right\}|$, which rapidly decays as the support $\sigma$ increases. However, thanks to early stopping, our implementation takes only a time proportional to $\min\{\,\alpha/\Psi(x_{S},n,N) ,|\left\{S \in \mathcal{S} | x_{S} \ge \sigma \right\}|\,\}$. Since $\alpha/\Psi(x_{S},n,N)$ is small for small values of $\sigma$, the iterations of our scheme are not slowed down by the potentially large number of frequent patterns when the support is near 1. In contrast, the original implementation of LAMP with decremental search needs $N-\sigma_{\mathrm{rt}}$ executions of frequent itemset mining to converge, whereas our scheme needs only $\sigma_{\mathrm{rt}}$. Therefore, the larger the ratio $(N-\sigma_{\mathrm{rt}})/\sigma_{\mathrm{rt}}$, the more advantage our proposal provides. As we can see in Table~\ref{table:datasets_summary}, in most cases the ratio is fairly large, which explains why incremental search with early stopping is empirically much faster than decremental search.

Next we study the effect of subsampling by applying our version of LAMP with subsampling ratios $K \in \left\{2,4,8,10,50,100\right\}$. 10 repetitions of the experiment were performed for each dataset; the average behaviour is depicted in Figure \ref{fig:subsampling}. We can confirm that the speed-up obtained by subsampling is overall proportional to $K$. This is shown in Figure  \ref{fig:subsampling} (I).

Since the minimum $\sigma_{\mathrm{rt}}$ which can be estimated from the subsampled datasets is equal to the subsampling ratio $K$, then the subsampling ratio should ideally be small compared to $\sigma_{\mathrm{rt}}$ if an accurate estimation of the number of testable hypotheses is needed.
As shown in Figure~\ref{fig:subsampling} (II), the estimated Bonferroni factor is indeed reasonably accurate for most datasets when a small subsampling ratio is used. Empirically we find that subsampling ratios of $K=2$ and $K=4$ are able to determine the number of testable hypotheses up to the right order of magnitude in the majority of cases. On the contrary, more aggressive subsampling ratios like $K \ge 10$, which provide speed ups of one of two orders of magnitude, are accurate only in some datasets. To sum up, when accurate solutions are needed, subsampling can only provide modest yet worthy speed ups on top of incremental search with early stopping, making the algorithm approximately up to 4 times faster.

\section{Conclusions}

In this work, we have proposed a fast algorithm for mining statistically significant higher-order interactions. We have shown empirically that by using a novel incremental search strategy with early stopping, the execution time with respect to the state-of-the-art approach \cite{Terada} is reduced by several orders of magnitude in the vast majority of the 11 datasets and 18 cases analysed. In practice, this brings down the expected computation time for real-world datasets from several days to just a few hours, making statistical testing of higher-order interactions feasible in more demanding datasets. Furthermore, for cases in which trading off accuracy for speed is deemed acceptable, we have explored the possibility of speeding up LAMP even further by subsampling datasets across observations. We have observed empirically that using moderate subsampling ratios can provide an extra speedup of up to one order of magnitude while retaining a reasonable accuracy in the determination of the number of testable hypotheses.

There are several aspects in which LAMP could be enhanced further. Firstly, since higher-order combinations have a hierarchical nature, the resulting test statistics will be highly correlated. Several schemes to correct for mutual dependence across tests in multiple hypothesis testing have been proposed both in a general context \cite{Nyholt04} and specific to binary predictors \cite{Zhang08}, \cite{Moskvina}. However, how to scale those approaches to datasets containing tens of thousands of features by integrating them into the LAMP framework remains an open problem. Another promising approach to reduce the number of testable hypotheses and gain statistical power is the integration of prior knowledge in the search procedure by exploiting the functional relations among predictors to prune the search space. Finally, extending the framework to deal with continuous labels, continuous features or even structured data represents another interesting way to extend the applicability of LAMP to new datasets and domains.

\clearpage
\clearpage

\renewcommand{\thefigure}{S\arabic{figure}}
\renewcommand{\thetable}{S\arabic{table}}

\appendix

\section{Dataset description}

The datasets T10I4D100K, T40I10D100K, retail, chess, connect, BMS-Web-2 and BMS-POS are well-known public benchmark datasets for frequent itemset mining \cite{FIMIDatasets}. Even though they are unlabeled datasets, we can still make use of them to evaluate how efficiently different search schemes enumerate all testable patterns.
Note that enumeration of testable patterns only requires to know the number $n$ of minority class examples, and the total number $N$ of examples. Moreover, the computational effort  needed to find the root frequency $\sigma_{rt}$ depends on the class labels only through the ratio\footnote{This can be seen by approximating $\Psi(\sigma,n,N)=\binom{n}{\sigma} / \binom{N}{\sigma} \approx (n/N)^{\sigma}$, which holds for $\sigma \ll n$ and is therefore a valid approximation for incremental search}$N/n$. Therefore, we can artificially fix a ratio $N/n$ and make use of those datasets to compare how efficiently different search schemes find $\sigma_{rt}$. In our experiments, we have considered both $N/n = 2$ and $N/n =10 $. The former can be seen as an optimistic assumption, as it will yield the biggest reduction in the Bonferroni Correction factor, while the latter represents rather unfavourable scenarios. 

Another 4 additional labeled datasets were used: tic-tac-toe\footnote{https://archive.ics.uci.edu/ml/datasets/Tic-Tac-Toe+Endgame}, internet advertisements\footnote{https://archive.ics.uci.edu/ml/datasets/Internet+Advertisements}, mushroom, and RCV1. The first three are well-known datasets from the UCI repository. Tic-tac-toe was binarized using dummy variables to represent the three possible states (empty, ``x'' or ``o'') of each space in the $3 \times 3$ grid; internet advertisements was kept as in the original, but 3 continuous features and 1 binary feature having missing values were discarded; mushroom was binarized as in \cite{FIMIDatasets} and the first two features, which are complementary and indicate whether the mushroom is edible or not, were used to define the labels. Finally, RCV1 is a well-known text classification benchmark. We used a reduced two-class version of the original dataset\footnote{\url{http://www.csie.ntu.edu.tw/~cjlin/libsvmtools/datasets/binary.html}, see \cite{Lewis2004} for a description of the original dataset}, and attributes were binarized depending on whether the \emph{tf-idf} of the corresponding word stem is positive or not as in \cite{RIT}.

\section{Statistical power of LAMP versus naive Bonferroni correction}

\begin{figure}[t]
	\begin{center}
		\includegraphics[width=1.0\textwidth]{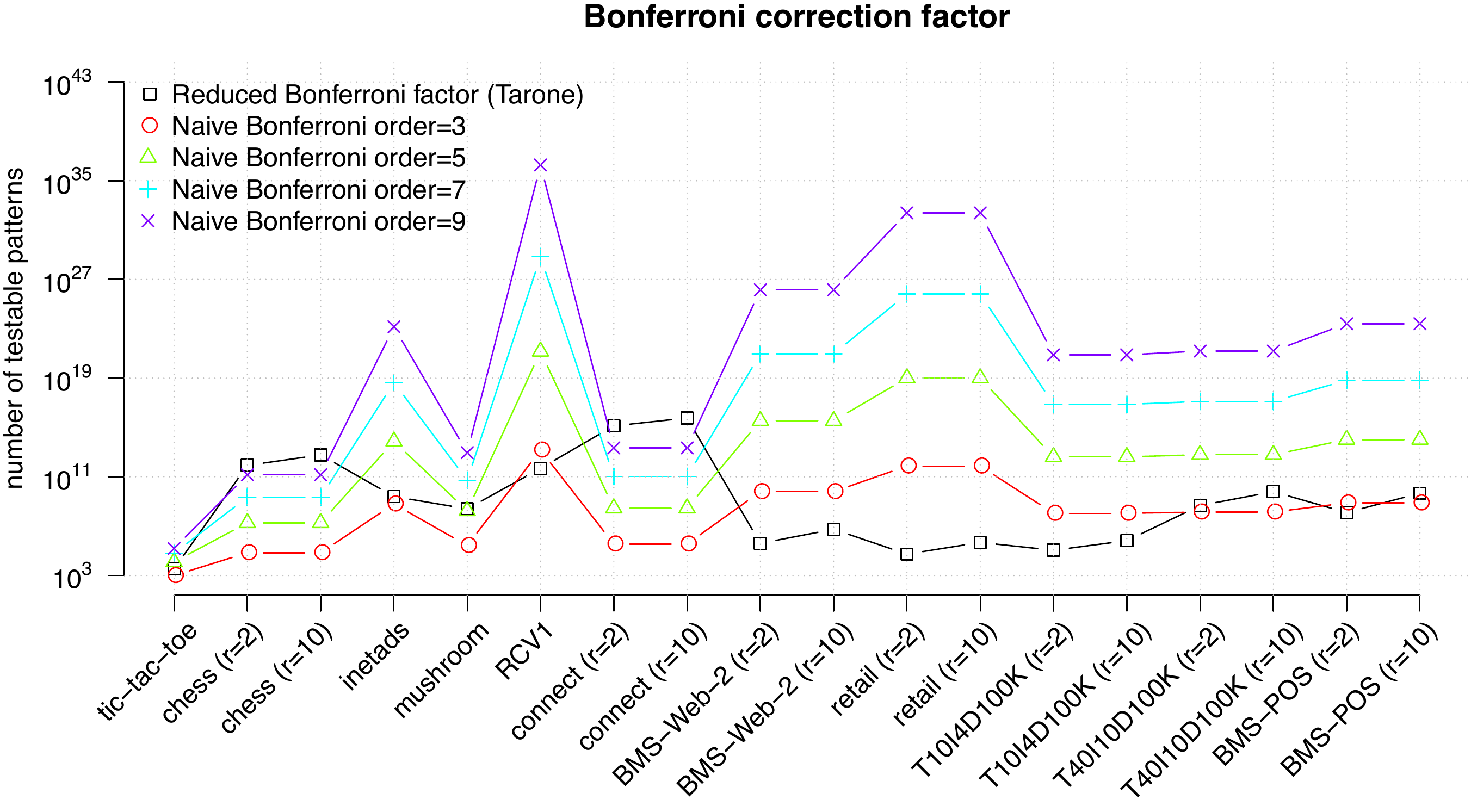}
	\end{center}
	\caption{Bonferroni correction factor obtained by LAMP when testing interactions of any order versus naive Bonferroni correction factors obtained by restricting the maximum order or the interactions to 3, 5, 7, and 9.}
	\label{fig:bonferroni}
\end{figure}

In Figure \ref{fig:bonferroni} we show the difference between the naive Bonferroni correction factor and the improved (reduced) Bonferroni factor first proposed by Tarone \cite{Tarone} and used by LAMP in different scenarios we have considered. Note that the correction factor used by LAMP effectively accounts for interactions of any order. On the contrary, the naive Bonferroni correction factor for testing interactions of any arity would equal $2^P$; an insurmountably big number for all datasets except Tic-tac-toe. Thus we restrict the order of interactions to 3, 5, 7, and 9 to make computation of the naive Bonferroni correction factor feasible.
Figure~\ref{fig:bonferroni} shows that even for a fixed arity, the number of hypotheses testing examined by the naive Bonferroni approach is much bigger, resulting in a great loss in statistical power.

\begin{sidewaystable}[htbp] 
		\caption{EXECUTION TIME} 
		\label{tab:results_ext1} 
		\centering 
			\begin{tabular}{c c c c c c c c c} 
			\hline 
			Dataset& LAMP & IncLAMP & $K=2$ & $K=4$ & $K=8$ & $K=10$ & $K=50$ & $K=100$ \\
			\hline 
tic-tac-toe & $2.97$ & $0.06$ & $0.13 \pm 0.21$ & $0.02 \pm 0.00$ & $0.08 \pm 0.04$ & $0.02 \pm 0.04$ & $0.01 \pm 0.00$ & $0.01 \pm 0.00$ \\
inetads & $14.99$ & $0.42$ & $0.15 \pm 0.20$ & $0.03 \pm 0.00$ & $0.07 \pm 0.01$ & $0.02 \pm 0.00$ & $0.01 \pm 0.00$ & $0.01 \pm 0.00$ \\
mushroom & $86.61$ & $0.91$ & $0.64 \pm 0.47$ & $0.19 \pm 0.01$ & $0.21 \pm 0.01$ & $0.15 \pm 0.01$ & $0.04 \pm 0.00$ & $0.02 \pm 0.00$ \\
RCV1 & $3876.88$ & $2701.23$ & $724.87 \pm 96.82$ & $401.02 \pm 115.00$ & $168.23 \pm 55.27$ & $129.78 \pm 32.88$ & $76.25 \pm 16.53$ & $56.86 \pm 29.05$ \\
T10I4D100K (r=2) & $1534.69$ & $3.69$ & $1.32 \pm 0.29$ & $0.52 \pm 0.02$ & $0.35 \pm 0.02$ & $0.20 \pm 0.00$ & $0.15 \pm 0.01$ & $0.06 \pm 0.00$ \\
T10I4D100K (r=10) & $686.08$ & $3.81$ & $1.75 \pm 0.13$ & $1.71 \pm 0.03$ & $1.10 \pm 0.02$ & $2.12 \pm 0.06$ & $0.15 \pm 0.01$ & $0.06 \pm 0.00$ \\
T40I10D100K (r=2) & $33250.01$ & $2475.01$ & $685.19 \pm 21.21$ & $465.00 \pm 15.92$ & $233.30 \pm 29.44$ & $154.03 \pm 11.85$ & $190.29 \pm 8.31$ & $63.23 \pm 3.83$ \\
T40I10D100K (r=10) & $76753.79$ & $8430.95$ & $1915.71 \pm 163.12$ & $1126.13 \pm 35.75$ & $1282.33 \pm 44.41$ & $799.22 \pm 28.62$ & $300.34 \pm 19.79$ & $62.17 \pm 2.95$ \\
retail (r=2) & $1980.96$ & $2.16$ & $1.00 \pm 0.30$ & $0.35 \pm 0.02$ & $0.28 \pm 0.03$ & $0.13 \pm 0.02$ & $0.07 \pm 0.00$ & $0.03 \pm 0.00$ \\
retail (r=10) & $437.82$ & $1.10$ & $0.73 \pm 0.16$ & $0.34 \pm 0.01$ & $0.28 \pm 0.01$ & $0.36 \pm 0.02$ & $0.07 \pm 0.00$ & $0.03 \pm 0.00$ \\
chess (r=2) & $21216.10$ & $765.84$ & $256.60 \pm 19.48$ & $99.80 \pm 7.73$ & $43.95 \pm 10.48$ & $28.11 \pm 5.90$ & $0.97 \pm 0.15$ & $0.18 \pm 0.05$ \\
chess (r=10) & $26423.52$ & $538.03$ & $347.15 \pm 29.93$ & $104.57 \pm 8.13$ & $36.65 \pm 8.39$ & $23.76 \pm 4.49$ & $0.97 \pm 0.14$ & $0.18 \pm 0.05$ \\
connect (r=2) & $526769.80$ & $4153.28$ & $2105.87 \pm 19.52$ & $834.20 \pm 12.95$ & $369.75 \pm 5.75$ & $353.22 \pm 32.76$ & $66.40 \pm 1.79$ & $23.19 \pm 1.15$ \\
connect (r=10) & $571678.15$ & $4429.63$ & $1550.34 \pm 30.22$ & $975.03 \pm 5.98$ & $554.27 \pm 3.71$ & $350.68 \pm 4.97$ & $56.43 \pm 1.35$ & $23.58 \pm 1.32$ \\
BMS-Web-2 (r=2) & $706.31$ & $1.29$ & $0.43 \pm 0.07$ & $0.15 \pm 0.00$ & $0.10 \pm 0.01$ & $0.10 \pm 0.00$ & $0.05 \pm 0.01$ & $0.03 \pm 0.01$ \\ 
BMS-POS (r=2) & $65646.16$ & $46.86$ & $16.87 \pm 0.61$ & $9.69 \pm 0.87$ & $8.08 \pm 1.03$ & $5.89 \pm 0.63$ & $4.00 \pm 0.49$ & $1.17 \pm 0.22$ \\
BMS-Web-2 (r=10) & $200.00$ & $1.19$ & $0.37 \pm 0.04$ & $0.19 \pm 0.01$ & $0.12 \pm 0.04$ & $0.17 \pm 0.01$ & $0.03 \pm 0.00$ & $0.02 \pm 0.00$ \\
BMS-POS (r=10) & $17433.12$ & $173.18$ & $60.01 \pm 2.75$ & $51.32 \pm 8.69$ & $27.55 \pm 1.97$ & $20.17 \pm 2.43$ & $3.93 \pm 0.48$ & $1.16 \pm 0.23$ \\[1ex]
			\hline 
			\end{tabular} 
\end{sidewaystable}

\begin{sidewaystable}[htbp] 
		\caption{ESTIMATED NUMBER OF TESTABLE PATTERNS ($K = 2$, $4$, and $8$)} 
		\label{tab:resultsext_2} 
		\centering 
			\begin{tabular}{c c c c c} 
			\hline 
			Dataset & Exact & $K=2$ & $K=4$ & $K=8$ \\
			\hline 
tic-tac-toe & $3.46e+03$ & $3.42e+03 \pm 1.74e+02$ & $3.86e+03 \pm 1.73e+02$ & $3.43e+03 \pm 3.60e+02$ \\
inetads & $2.52e+09$ & $1.67e+08 \pm 1.91e+07$ & $2.12e+08 \pm 4.79e+07$ & $2.47e+08 \pm 3.51e+07$\\
mushroom & $2.52e+08$ & $2.37e+08 \pm 1.01e+06$ & $2.60e+08 \pm 1.13e+06$ & $3.08e+08 \pm 2.84e+06$ \\
RCV1 & $4.82e+11$ & $8.36e+11 \pm 7.15e+11$ & $1.11e+12 \pm 1.22e+12$ & $7.79e+12 \pm 8.61e+12$ \\
T10I4D100K (r=2) & $1.11e+05$ & $1.23e+05 \pm 3.82e+03$ & $1.32e+05 \pm 2.89e+03$ & $5.41e+05 \pm 7.38e+04$ \\
T10I4D100K (r=10) & $6.66e+05$ & $2.64e+06 \pm 2.53e+05$ & $2.60e+06 \pm 1.40e+05$ & $6.95e+06 \pm 4.26e+05$\\
T40I10D100K (r=2) & $4.68e+08$ & $2.77e+08 \pm 4.30e+06$ & $3.01e+08 \pm 9.09e+06$ & $4.04e+08 \pm 3.78e+07$\\
T40I10D100K (r=10) & $6.22e+09$ & $7.68e+09 \pm 4.11e+09$ & $4.37e+12 \pm 1.02e+12$ & $1.69e+14 \pm 7.19e+12$\\
retail (r=2) & $5.39e+04$ & $5.53e+04 \pm 6.65e+02$ & $5.93e+04 \pm 1.63e+03$ & $5.93e+04 \pm 2.09e+04$\\
retail (r=10) & $4.81e+05$ & $7.39e+07 \pm 1.21e+08$ & $7.51e+09 \pm 2.62e+09$ & $2.64e+10 \pm 6.56e+09$\\
chess (r=2) & $8.39e+11$ & $9.26e+11 \pm 1.74e+10$ & $1.06e+12 \pm 3.50e+10$ & $1.08e+12 \pm 6.34e+10$\\ 
chess (r=10) & $5.96e+12$ & $5.73e+12 \pm 3.84e+10$ & $7.01e+12 \pm 9.45e+10$ & $9.55e+12 \pm 9.74e+10$\\
connect (r=2) & $1.33e+15$ & $1.36e+15 \pm 1.65e+12$ & $1.44e+15 \pm 3.02e+12$ & $1.61e+15 \pm 3.63e+12$\\
connect (r=10) & $5.86e+15$ & $6.29e+15 \pm 3.01e+12$ & $5.76e+15 \pm 4.99e+12$ & $4.62e+15 \pm 8.53e+12$\\
BMS-Web-2 (r=2) & $4.02e+05$ & $1.80e+05 \pm 1.89e+04$ & $2.03e+05 \pm 3.83e+04$ & $2.25e+05 \pm 1.45e+05$\\
BMS-POS (r=2) & $1.26e+08$ & $1.01e+08 \pm 1.23e+07$ & $1.55e+08 \pm 2.96e+07$ & $1.93e+08 \pm 6.85e+07$\\
BMS-Web-2 (r=10) & $5.60e+06$ & $1.91e+06 \pm 9.01e+05$ & $7.26e+06 \pm 2.14e+06$ & $2.14e+07 \pm 4.64e+06$ \\
BMS-POS (r=10) & $4.68e+09$ & $6.53e+09 \pm 7.28e+09$ & $4.99e+11 \pm 1.04e+12$ & $1.04e+13 \pm 3.35e+12$ \\[1ex]
			\hline 
			\end{tabular}
\end{sidewaystable}

\begin{sidewaystable}[htbp] 
		\caption{ESTIMATED NUMBER OF TESTABLE PATTERNS ($K = 10$, $50$, and $100$)} 
		\label{tab:resultsext_2} 
		\centering 
			\begin{tabular}{c c c c c} 
			\hline 
			Dataset & Exact & $K=10$ & $K=50$ & $K=100$ \\
			\hline 
tic-tac-toe & $3.46e+03$ & $2.77e+03 \pm 2.61e+02$ & $1.95e+03 \pm 3.76e+02$ & $1.14e+03 \pm 1.68e+02$ \\
inetads & $2.52e+09$ & $1.82e+08 \pm 2.87e+07$ & $2.66e+08 \pm 9.44e+07$ & $1.19e+08 \pm 3.28e+07$ \\
mushroom & $2.52e+08$ & $2.35e+08 \pm 2.74e+06$ & $4.70e+08 \pm 4.40e+06$ & $2.71e+08 \pm 6.55e+06$ \\
RCV1 & $4.82e+11$ & $3.53e+13 \pm 6.38e+13$ & $1.84e+14 \pm 2.00e+14$ & $9.99e+14 \pm 4.80e+14$ \\
T10I4D100K (r=2) & $1.11e+05$ & $3.15e+05 \pm 8.48e+04$ & $1.71e+07 \pm 1.18e+06$ & $8.84e+06 \pm 6.38e+05$ \\
T10I4D100K (r=10) & $6.66e+05$ & $8.12e+07 \pm 1.41e+06$ & $1.71e+07 \pm 1.18e+06$ & $8.84e+06 \pm 6.38e+05$ \\
T40I10D100K (r=2) & $4.68e+08$ & $1.01e+09 \pm 6.99e+08$ & $2.38e+12 \pm 6.75e+11$ & $7.62e+13 \pm 2.42e+12$ \\
T40I10D100K (r=10) & $6.22e+09$ & $1.01e+14 \pm 6.54e+12$ & $2.08e+14 \pm 7.01e+12$ & $7.62e+13 \pm 2.42e+12$ \\
retail (r=2) & $5.39e+04$ & $5.19e+06 \pm 1.14e+07$ & $7.52e+10 \pm 6.69e+09$ & $3.42e+10 \pm 4.54e+09$ \\
retail (r=10) & $4.81e+05$ & $1.54e+10 \pm 2.20e+09$ & $7.52e+10 \pm 6.69e+09$ & $3.42e+10 \pm 4.54e+09$ \\
chess (r=2) & $8.39e+11$ & $1.14e+12 \pm 8.56e+10$ & $3.42e+12 \pm 2.27e+11$ & $1.28e+12 \pm 1.08e+11$ \\ 
chess (r=10) & $5.96e+12$ & $7.47e+12 \pm 8.47e+10$ & $3.42e+12 \pm 2.27e+11$ & $1.28e+12 \pm 1.08e+11$ \\
connect (r=2) & $1.33e+15$ & $1.51e+15 \pm 2.97e+12$ & $8.19e+14 \pm 1.75e+12$ & $6.20e+14 \pm 5.24e+12$ \\
connect (r=10) & $5.86e+15$ & $5.09e+15 \pm 1.76e+13$ & $1.42e+15 \pm 8.34e+12$ & $6.20e+14 \pm 5.24e+12$ \\
BMS-Web-2 (r=2) & $4.02e+05$ & $6.41e+05 \pm 5.02e+05$ & $5.32e+07 \pm 7.33e+06$ & $2.88e+07 \pm 3.95e+06$ \\
BMS-POS (r=2) & $1.26e+08$ & $2.70e+08 \pm 1.29e+08$ & $1.22e+13 \pm 2.30e+12$ & $4.09e+12 \pm 1.48e+12$ \\
BMS-Web-2 (r=10) & $5.60e+06$ & $2.65e+08 \pm 1.65e+07$ & $5.32e+07 \pm 7.33e+06$ & $2.88e+07 \pm 3.95e+06$ \\
BMS-POS (r=10) & $4.68e+09$ & $5.88e+12 \pm 3.64e+12$ & $1.22e+13 \pm 2.30e+12$ & $4.09e+12 \pm 1.48e+12$ \\[1ex]
			\hline 
			\end{tabular}
\end{sidewaystable}

\begin{sidewaystable}[htbp] 
		\caption{ESTIMATED ROOT FREQUENCY} 
		\label{tab:results_ext3} 
		\centering 
			\begin{tabular}{c c c c c c c c} 
			\hline 
			Dataset & Exact & $K=2$ & $K=4$ & $K=8$ & $K=10$ & $K=50$ & $K=100$ \\
			\hline 
tic-tac-toe & $11.00$ & $12.00 \pm 0.00$ & $12.00 \pm 0.00$ & $16.00 \pm 0.00$ & $20.00 \pm 0.00$ & $50.00 \pm 0.00$ & $100.00 \pm 0.00$ \\
inetads & $13.00$ & $12.00 \pm 0.00$ & $12.00 \pm 0.00$ & $16.00 \pm 0.00$ & $20.00 \pm 0.00$ & $50.00 \pm 0.00$ & $100.00 \pm 0.00$ \\
mushroom & $31.00$ & $32.00 \pm 0.00$ & $32.00 \pm 0.00$ & $32.00 \pm 0.00$ & $40.00 \pm 0.00$ & $50.00 \pm 0.00$ & $100.00 \pm 0.00$ \\
RCV1 & $41.00$ & $42.80 \pm 1.40$ & $44.00 \pm 0.00$ & $48.00 \pm 0.00$ & $51.00 \pm 3.16$ & $100.00 \pm 0.00$ & $100.00 \pm 0.00$ \\
T10I4D100K (r=2) & $22.00$ & $22.00 \pm 0.00$ & $24.00 \pm 0.00$ & $24.00 \pm 0.00$ & $30.00 \pm 0.00$ & $50.00 \pm 0.00$ & $100.00 \pm 0.00$ \\
T10I4D100K (r=10) & $8.00$ & $8.00 \pm 0.00$ & $12.00 \pm 0.00$ & $16.00 \pm 0.00$ & $10.00 \pm 0.00$ & $50.00 \pm 0.00$ & $100.00 \pm 0.00$ \\
T40I10D100K (r=2) & $34.00$ & $34.00 \pm 0.00$ & $36.00 \pm 0.00$ & $40.00 \pm 0.00$ & $40.00 \pm 0.00$ & $100.00 \pm 0.00$ & $100.00 \pm 0.00$ \\
T40I10D100K (r=10) & $12.00$ & $13.40 \pm 0.97$ & $16.00 \pm 0.00$ & $16.00 \pm 0.00$ & $20.00 \pm 0.00$ & $50.00 \pm 0.00$ & $100.00 \pm 0.00$ \\
retail (r=2) & $21.00$ & $22.00 \pm 0.00$ & $24.00 \pm 0.00$ & $32.00 \pm 0.00$ & $37.00 \pm 4.83$ & $50.00 \pm 0.00$ & $100.00 \pm 0.00$ \\
retail (r=10) & $7.00$ & $10.20 \pm 0.63$ & $12.00 \pm 0.00$ & $16.00 \pm 0.00$ & $20.00 \pm 0.00$ & $50.00 \pm 0.00$ & $100.00 \pm 0.00$ \\
chess (r=2) & $44.00$ & $44.00 \pm 0.00$ & $44.00 \pm 0.00$ & $48.00 \pm 0.00$ & $50.00 \pm 0.00$ & $50.00 \pm 0.00$ & $100.00 \pm 0.00$ \\
chess (r=10) & $14.00$ & $16.00 \pm 0.00$ & $16.00 \pm 0.00$ & $16.00 \pm 0.00$ & $20.00 \pm 0.00$ & $50.00 \pm 0.00$ & $100.00 \pm 0.00$ \\
connect (r=2) & $55.00$ & $56.00 \pm 0.00$ & $56.00 \pm 0.00$ & $56.00 \pm 0.00$ & $60.00 \pm 0.00$ & $100.00 \pm 0.00$ & $100.00 \pm 0.00$ \\
connect (r=10) & $18.00$ & $18.00 \pm 0.00$ & $20.00 \pm 0.00$ & $24.00 \pm 0.00$ & $20.00 \pm 0.00$ & $50.00 \pm 0.00$ & $100.00 \pm 0.00$ \\
BMS-Web-2 (r=2) & $23.00$ & $22.60 \pm 0.97$ & $24.00 \pm 0.00$ & $31.20 \pm 2.53$ & $30.00 \pm 0.00$ & $50.00 \pm 0.00$ & $100.00 \pm 0.00$ \\
BMS-POS (r=2) & $32.00$ & $32.00 \pm 0.00$ & $32.00 \pm 0.00$ & $39.20 \pm 2.53$ & $40.00 \pm 0.00$ & $50.00 \pm 0.00$ & $100.00 \pm 0.00$ \\
BMS-Web-2 (r=10) & $9.00$ & $9.80 \pm 0.63$ & $12.00 \pm 0.00$ & $16.00 \pm 0.00$ & $10.00 \pm 0.00$ & $50.00 \pm 0.00$ & $100.00 \pm 0.00$ \\
BMS-POS (r=10) & $11.00$ & $12.00 \pm 0.00$ & $16.00 \pm 0.00$ & $16.00 \pm 0.00$ & $20.00 \pm 0.00$ & $50.00 \pm 0.00$ & $100.00 \pm 0.00$ \\[1ex]
			\hline 
			\end{tabular} 
\end{sidewaystable}

\clearpage

\bibliography{bibliography}

\begin{thebibliography}{10}

\bibitem{Agrawal93}
R.~Agrawal, T.~Imieli\'{n}ski, and A.~Swami.
\newblock Mining association rules between sets of items in large databases.
\newblock {\em SIGMOD Record}, 22(2):207--216, 1993.

\bibitem{Toivonen}
H.~Toivonen.
\newblock Sampling large databases for association rules.
\newblock In {\em Proceedings of the 22th International Conference on Very
  Large Data Bases}, pages 134--145, 1996.

\bibitem{RIT}
R.~D. Shah and N.~Meinshausen.
\newblock Random intersection trees.
\newblock {\em Journal of Machine Learning Research}, 15:629--654, 2014.

\bibitem{Terada}
A.~Terada, M.~Okada-Hatakeyama, K.~Tsuda, and J.~Sese.
\newblock Statistical significance of combinatorial regulations.
\newblock {\em Proceedings of the National Academy of Sciences}, 2013.

\bibitem{Tarone}
R.~E. Tarone.
\newblock A modified bonferroni method for discrete data.
\newblock {\em Biometrics}, 46(2):515--22, 1990.

\bibitem{FisherExactTest}
R.~A. Fisher.
\newblock On the interpretation of $\chi^{2}$ from contingency tables, and the
  calculation of \emph{P}.
\newblock {\em Journal of the Royal Statistical Society}, 85(1):87--94, 1922.

\bibitem{bland2000}
J.~M. Bland.
\newblock {\em An introduction to medical statistics}.
\newblock Oxford University Press, 2000.

\bibitem{Bonferroni36}
C.~E. Bonferroni.
\newblock Teoria statistica delle classi e calcolo delle probabilit\`{a}.
\newblock {\em Pubblicazioni del R Istituto Superiore di Scienze Economiche e
  Commerciali di Firenze}, 8:3--62, 1936.

\bibitem{Uno04anefficient}
T.~Uno, T.~Asai, Y.~Uchida, and H.~Arimura.
\newblock An efficient algorithm for enumerating closed patterns in transaction
  databases.
\newblock In {\em Discovery Science}, volume 3245 of {\em LNCS}, pages 16--31,
  2004.

\bibitem{FIMIDatasets}
B.~Goethals and M.~J. Zaki.
\newblock Frequent itemset mining dataset repository (fimi' 04).
\newblock \url{http://fimi.ua.ac.be/data/}, 2004.

\bibitem{Nyholt04}
D.~R. Nyholt.
\newblock A simple correction for multiple testing for single-nucleotide
  polymorphisms in linkage disequilibrium with each other.
\newblock {\em The American Journal of Human Genetics}, 74(4):765--769, 2004.

\bibitem{Zhang08}
X.~Zhang, F.~Pan, W.~Wang, and A.~Nobel.
\newblock Mining non-redundant high order correlations in binary data.
\newblock {\em Proceedings of the VLDB Endowment}, 1(1):1178--1188, 2008.

\bibitem{Moskvina}
V.~Moskvina and K.~M. Schmidt.
\newblock On multiple-testing correction in genome-wide association studies.
\newblock {\em Genetic Epidemiology}, 32(6):567--573, 2008.

\bibitem{Lewis2004}
D.~D. Lewis, Y.~Yang, T.~G. Rose, and F.~Li.
\newblock {RCV1}: A new benchmark collection for text categorization research.
\newblock {\em Journal of Machine Learning Ressearch}, 5:361--397, 2004.

\end{thebibliography}
\bibliographystyle{unsrt}

\clearpage

\end{document}